\documentclass[5p]{elsarticle}
\usepackage{lineno,hyperref}
\journal{Computer Vision and Image Understanding}
\usepackage{amssymb}
\usepackage[figuresright]{rotating}
\usepackage{times}
\usepackage{epsfig}
\usepackage{graphicx}
\usepackage{amsmath}
\usepackage{amssymb}
\usepackage{lineno}
\usepackage{color}
\usepackage{xcolor}
\usepackage{blindtext}

\newcommand{\xx}{\mbox{x}}

\newcommand{\xxs}{\mbox{x}_S}
\newcommand{\xxt}{\mbox{x}_T}
\newcommand{\RR}{\mathbb{R}}
\newcommand{\ie}{\mbox{i.e.}}

\begin{document}

\begin{frontmatter}

\title{Location Recognition Over Large Time Lags}

\author{Basura 
Fernando\corref{mycorrespondingauthor}}
\cortext[mycorrespondingauthor]{Corresponding author}
\ead{basura.fernando@esat.kuleuven.be}
\fntext[myfootnote]{Telephone : +32163 72422}

\author{Tatiana Tommasi}
\author{Tinne Tuytelaars}

\address{KU Leuven ESAT-PSI, iMinds\\
Kasteelpark Arenberg 10 - bus 2441\\
B-3001 Heverlee, Belgium}

\begin{abstract}
Would it be possible to automatically associate ancient pictures to modern ones 
and create fancy cultural heritage city maps? 
We introduce here the task of recognizing the location depicted in an old photo 
given modern annotated images collected from the Internet. We present an extensive 
analysis on different features, looking for the most discriminative and most robust 
to the image variability induced by large time lags. 
Moreover, we show that the described task benefits from domain adaptation.
\end{abstract}


\end{frontmatter}


\section{Introduction}
\label{sec:intro}

A hundred year old photograph or a postcard can reveal a lot about our 
culture and history. Following this idea, many cultural heritage campaigns 
recently started to promote the digitization of large amounts of visual data. 
Several cities and towns all over the world, as well as institutions such as universities or museums, 
are bringing archives with their images and footage online, providing 
public access and calling for methods to efficiently open up and exploit these resources 
\cite{VTM,streetmuseum}.

\begin{figure}[t]
\centering
\includegraphics[width=0.9\linewidth]{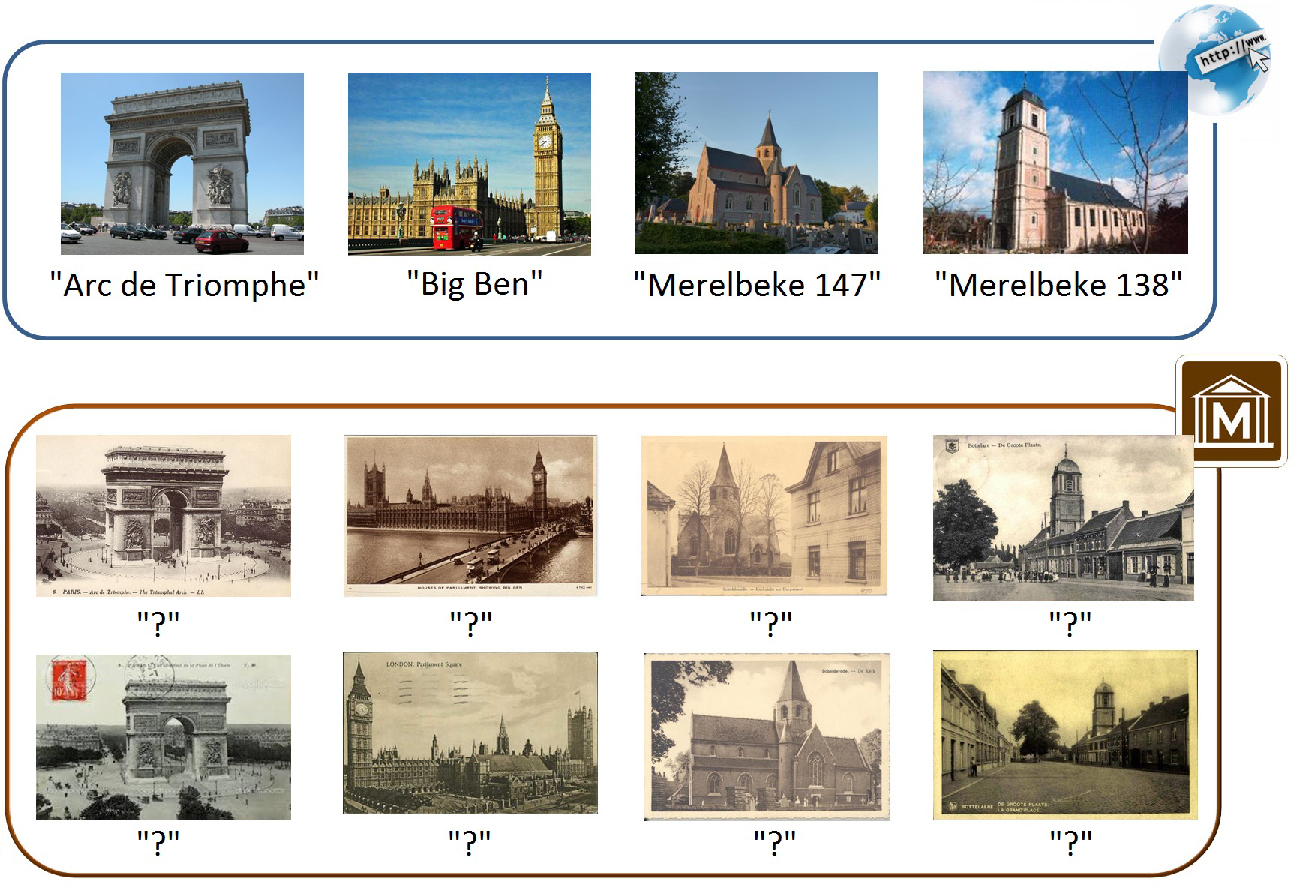}
\caption{Pictures of four locations over large time lags showing an evident
change in visual appearance. 
The photographs are similar in their high
level scene content, but the color range and texture are significantly
different. Modern photos can be easily found on the World Wide Web, while
ancient pictures are provided by cultural heritage museums.  The task we
address in this paper consists in annotating ancient pictures 
given a set of labeled modern images.
}
\label{fig:newvsold}
\end{figure}

At the time when photography was not affordable for private and everyday use, 
most of the pictures were taken in public places and depict buildings, monuments,
statues, or more in general, common locations of interest. Some of those are
landmarks and tourist attractions. Others are locations with 
historical value. Popular landmarks often appear in modern digital images which 
are shared online through applications such as Flickr. Other historical 
locations can be associated to their geographic coordinates through 
Google Maps and visualized by means of applications like Google Street-View.
Despite the place correspondence, the visual appearance of old and new images 
is dramatically different.
As shown in Figure~\ref{fig:newvsold}, ancient photographs have different colors, texture, 
and contrast characteristics compared to modern digital images \cite{PalermoHE12}. 
Moreover it is not possible to control the acquisition perspective: 
changes in the urban planning along the years may have made some viewpoints not accessible.

Numerous efforts have been dedicated to recognizing landmarks in image databases containing 
photographs of the same era~\cite{Li2010,Gronat2013,Cao2013,Torii2013},
but to our knowledge, no previous work focused on tackling location recognition over large time lags. 
Here we define this task: \textbf{annotate an ancient photograph with the correct location label, given a 
set of labeled modern photos}. In particular, we propose several useful tools to 
cope with this problem, making three main contributions:
\begin{list}{\labelitemi}{\leftmargin=1em}
\item  
we introduce a collection of images spanning over 25 locations and more
than one century, with the eldest photographs dating back to the 1850s; 
\item  
we present a detailed analysis of existing feature representations,
looking for the most robust features, suitable to handle the variability 
induced by different imaging processes adopted over time;
\item 
old and new images can be considered as belonging to two different domains.
We use existing domain adaptation methods and we show promising 
results in both location recognition and interactive location retrieval.
\end{list}

The rest of the paper is organized as follows.
Section \ref{sec:rel} revises the related work on location recognition and domain adaptation.
Section \ref{sec:chall} introduces our Large Time Lags Locations dataset and
indicates the challenges of location recognition on this testbed.
Section \ref{sec:da} briefly reviews the domain adaptation methods used in our study.
In section \ref{sec:expers} we present and discuss the obtained experimental results.
Finally, section \ref{sec:concl} concludes the paper and points out possible directions for
future research.

\section{Related Work}
\label{sec:rel}

\emph{Location recognition} consists in determining where a photo was taken 
by using as reference a database of previously seen locations \cite{Li2010}.
The interest towards this task grew together with the number of freely available
images on the Internet, many of which are geo-tagged and depict urban outdoor
scenes. Today, with the widespread use of mobile devices endowed with built-in cameras
and Internet connectivity, location recognition is a useful tool
for city guides and smart navigation aids that are able to localize an image in 
near real time \cite{googleglassWACV2014, survey2009}.

Given a structured database covering a pre-defined set of places, location recognition 
can be tackled as a classification problem \cite{Gronat2013,Cao2013}. 
The models for each place are learned offline and, at query time, a photograph 
is localized by assigning to it the label of the best scoring location classifier \cite{Gronat2013} .
Previous work also considered this task as a retrieval problem: a query image is 
used to find a set of similar images from a database which are then returned as place 
suggestions \cite{Torii2013,Schindler2007,Li2008}. This setting is mainly adopted when 
dealing with reference image collections possibly containing a large number of distractors.

Regardless of the chosen setup, one of the main challenges for location recognition
is the choice of appropriate image descriptors. The variability in illumination conditions, 
viewpoint and occlusion can dramatically influence the similarity of images even depicting 
the same place or building. 
The data similarity is generally based on local descriptors and Bag-Of-Words (BOW) based 
techniques \cite{SivicBOW}, and the retrieval is performed by computing distances between sparse
BOW histograms \cite{Philbin07}. Several improvements on this core system have 
been proposed by learning better descriptors \cite{doersch2012what,Philbin2010}, introducing 
more accurate descriptor matching \cite{Jegou2008}, exploiting 3D point clouds as powerful
representations \cite{Li2010,Sattler2011}, or carefully handling repetitive structures such 
as building facades \cite{Torii2013}. 

The mentioned large visual variability occurs in spite of the standard practice of using
photos acquired with high resolution modern cameras for location recognition. Although urban 
scenes and landmarks have been often captured even in ancient pictures and paintings, these samples are
generally neglected and the further issues induced by vintage color processes or artistic
brushstrokes are not considered in this task in the literature. 
One attempt to define robust detectors and descriptors was presented in \cite{Bansal2013,Hauagge2012}, 
where local symmetry features and 
spectral correspondence methods are proposed to match urban scenes with lighting, age and 
rendering style variations. The problems of alignment between paintings and photographs \cite{M.Aubry2013,Russell2011}
and viewpoint re-capturing over time \cite{Bae2010} have been tackled mainly leveraging over 3D models.
The pioneering work of Shrivastava et al. ~\cite{Shrivastava2011} defined visual 
similarities between paintings and pictures taken in different seasons. The 
proposed method relies on the robustness of HOG features \cite{HOG} and leverages the 
visual uniqueness of query images against millions of negative data. 
Despite their relevance, all these approaches have not been tested before for location 
recognition.

Solving the problem induced by data variability is also one of the goals of \emph{domain adaptation} \cite{DAsurvey}.
Instead of focusing directly on image-pairs matching, domain adaptation examines the
data distributions from which the images are drawn. 
Specifically, two sets of data are considered as belonging to two different domains if 
they cover the same set of classes but their marginal 
distributions differ. The aim of domain adaptation is to reduce this distribution shift \cite{DAsurvey}.
Various approaches fulfill this purpose by sample re-weighting
and selection \cite{lst_nips11,Gong2013}, self-labeling \cite{dasvm,Tommasi2013} and 
metric learning \cite{Kulis2011,Saenko2010}. A solution that has recently received a lot
of attention in the computer vision community consists in embedding the samples in a low
dimensional subspace shared by both the domains and invariant to their specific characteristics \cite{Fernando2013,Fernando2014,Gong2012,Gopalan2011}.
This strategy allows to tackle cases where the samples present originally high dimensional 
feature vectors and one of the two domains contains only unlabeled samples (unsupervised domain adaptation).

Previous work demonstrated that time can naturally cause a visual domain shift \cite{Konstantinos2013a,Hoffman2014}.
Existing methods applied to close this time gap proposed to discover object-specific style-sensitive patches \cite{Lee_2013},
to predict the behavior of time-varying probability distributions \cite{CLampert_arxiv2014} or to learn models adaptively over
a continuous manifold \cite{Hoffman2014}. However, all these approaches require details about the time ordering
(evolution) of images, which is often difficult to obtain, especially with ancient photographs. 
In many cases only two set of data are available, one older than the other without any further information. 
Our work fits in this context. 
We focus on the problem of location recognition over large time lags where we are given
a set of labeled modern photos and we want to annotate unlabeled historical pictures.

\section{The Large Time Lags Locations Dataset}
\label{sec:chall}
As detailed earlier, location recognition has so far been studied over modern 
images and the issues induced by large time lags have been only marginally 
considered for other tasks. Therefore one of the contributions of this paper is a
database of images which spans over a wide time period and numerous locations. The dataset
is presented in this section and used throughout the paper.

\subsection{Details of the dataset}

\begin{table}[t]
\centering
\begin{tabular}{|c|c|c|c|} \hline
Image Set & minimum & maximum & mean \\ \hline
New Images & 4 & 22 & 11 \\ \hline
Old Images & 1 & 22 & 8 \\ \hline
Dataset & 6 & 36 & 19 \\ \hline
\end{tabular}
\caption{Some dataset statistics. Minimum, maximum and mean number of images per class is shown.}
\label{tbl:stats}
\end{table}

We introduce here our Large Time Lags Locations (LTLL) dataset containing pictures of 25 
locations captured over a range of more than 150 years. Specifically, we collected images 
from several cities and towns in Europe such as Paris, London, Merelbeke, Leuven and ancient 
cities from Asia such as Agra in India,  Colombo and Kandy from Sri Lanka. We chose thirteen 
locations considering the presence of well known landmarks for which it has been easy to 
download old and new pictures from the Web. The remaining twelve locations are in the 
municipality of Merelbeke, Flemish Province of East Flanders in Belgium. Ancient 
images of these historical locations dating back to the period 1850s-1950s have been provided 
by the city archive of Merelbeke. We downloaded the 
corresponding modern images from Flickr, Google Street-View and the Google-Images search engine, 
although for some of the locations only a limited amount of modern photos could be obtained. Some
statistics about the dataset is shown in Table~\ref{tbl:stats}.

In total the dataset contains 225 historical pictures and 275 modern ones. More details on the 
images and their metadata are available from our project 
web-page\footnote{\url{http://homes.esat.kuleuven.be/~bfernand/beeldcanon/}}.

\subsection{Goals and Challenges}

Our main goal is to recognize the location of an old picture using annotated 
modern photographs. Primarily, location recognition in this setting can be considered as an image 
classification task. In this paper we use the LTLL dataset to investigate the effectiveness of 
existing location recognition tools, following the most typical image classification framework
and using the standard pipeline with feature detection, description and encoding \cite{Chatfield2011}.
In comparison to previous location recognition benchmarks, the LTLL dataset poses new challenges
related to the fact that the photos come from two different eras and to the limited amount of reference 
modern images for some historical place of cultural interest.

Given the LTLL dataset as testbed, we want to establish which of the existing 
feature detectors (Difference of Gaussians (DoG~\cite{marr80}), Hessian 
Affine~\cite{Perdoch2009}, etc.), feature descriptors (SIFT, LIOP~\cite{Wang2011a}, etc.) 
and representations ( BOW, Fisher Vectors \cite{Perronnin2010}, DeCAF~\cite{Donahue2014}) 
is able to cope better with the image variability due to large time lags. 

Due to variations in the capturing process as well as image degradation, old and new photographs 
belong to two different data distributions. Machine learning adaptive techniques are generally
used in classification to overcome this kind of distribution mismatch issues.
We investigate whether domain adaptation can help in reducing the distribution shift
between old and new photographs in the LTLL database. We start our analysis by adopting 
a classification setup with the modern images as training set (source) and the historical 
images as test samples (target). Apart from using all the images at once we also evaluate 
empirically the problems induced by the lack of modern data in the extreme case of having from 
one to five available training samples per location. 

Finally, by combining the LTLL database with a large set of modern image distractors, 
we extend our study to cross-domain location retrieval.  Here the ancient 
images are used as queries and the modern photos constitute the reference archive. 

Before going into the details of the experimental analysis (provided in section \ref{sec:expers}),
we dedicate the next section to a brief review of the considered domain adaptation methods.

\section{Subspace Domain Adaptation}
\label{sec:da}

Among the existing domain adaptation approaches, we consider here three methods 
based on subspace learning. Most of the location recognition solutions rely on high 
dimensional features such as HOG or BOW with large vocabulary dimension of $10^3-10^6$  
words (see e.g. \cite{Gronat2013,Cao2013}), and Fisher Vectors (FV, \cite{Perronnin2010,Jaakkola98}). 
Thus, using dimensionality reduction techniques appears to be a viable option. 
In the following we review the Geodesic Flow Kernel (GFK) method \cite{Gong2012} and the 
Subspace Alignment (SA) approach \cite{Fernando2013} together with its Extended (ESA) 
version presented in \cite{Fernando2014}. 
All these domain adaptation methods are unsupervised: they operate directly on the data 
representation with the labels available only for the source domain. 
In the following subsections we specify the differences among them and the various 
strategies used to estimate the subspace dimensionality. 

\vspace{2mm}
Let's indicate with ~$\xxs,\xxt \in \RR ^{1 \times D}$ the samples belonging respectively 
to a \emph{source} (training data, in our case new images which are labeled) and a \emph{target} 
(testing data, in our case old images) domain. We assume to obtain the source domain subspace 
$X_S\in \RR ^{D \times d_S}$, and the target domain subspace $X_T\in \RR ^{D \times d_T}$ 
by PCA, where $d_S,d_T<D$~ correspond to the number of selected eigenvectors associated 
with the largest eigenvalues. 

\subsection{GFK: Geodesic Flow Kernel} 

The GFK technique fixes the same dimensionality $d = d_S=d_T$ for the subspaces of the two 
domains and embeds them onto a Grassmann manifold. The geodesic flow $\{\Phi(t): t \in [0,1]\}$ 
between $X_S=\Phi(0)$  and $X_T=\Phi(1)$ is then used to parametrize the connection among the 
subspaces and to define infinitely many features varying gradually from the source to the target 
~$z^\infty=\{\Phi(t)^\top\xx: t \in [0,1]\}$. The inner product of the new features gives rise
to a positive semidefinite kernel \cite{Gong2012}
\begin{equation}
 Sim(\xx_i, \xx_j) = \langle z^\infty_i,z^\infty_j\rangle = \xx_i^\top\int_0^1 \Phi(t)\Phi(t)^\top dt ~\xx_j = \xx_i \mathbf{G} \xx_j~,
 \label{eq:gfk}
\end{equation}
where the matrix $\mathbf{G}$ can be calculated efficiently using singular value decomposition. 
The sample similarity obtained in this way is far less sensitive to the original domain differences. 
The dimensionality $d$ is chosen by optimizing a \emph{subspace disagreement measure} (SDM) 
that evaluates the similarity among the source, the target and the combined source+target subspace. 
For more details, we refer to \cite{Gong2012}. 

\subsection{SA: Subspace Alignment} 

The SA method  learns a linear transformation matrix ~$M\in \RR ^{d_S \times d_T}$~ that aligns the 
source and target coordinate systems by minimizing the following Bregman divergence: 
\begin{equation}
F(M) = || X_S M - X_T ||_F^{2}~,
\label{eq:objective1}
\end{equation}
where $||.||_F^{2}$ is the Frobenius norm. It can be easily shown that the optimal matrix is 
$M  = X_S' X_T$~, and the \textit{target aligned source coordinate system} is $X_a = X_S X_S' X_T$~. 
Finally, the similarity among two samples is defined as follows:
\begin{equation}
 Sim(\xxs, \xxt) = (\xxs X_a)(\xxt X_T)'~.
\label{eq:sim}
\end{equation}
It is possible to demonstrate that the deviation between two successive eigenvalues is bounded 
\cite{Fernando2013}. The bound can be used to determine the maximum size of the 
subspaces $d_{max}$  that allows to get a stable and non overfitting matrix $M$. The choice of 
the subspace dimensionality $d$ can then be done by minimizing the classification error through
a two fold \emph{cross-validation} over the labeled source data and finally setting $d_S=d_T=d$. 
For more details, we refer the reader to \cite{Fernando2013}.

\subsection{ESA: Extended Subspace Alignment} 
The function in (\ref{eq:sim}) operates in the original $\RR ^D$ space. However, after the domain 
transformation any problem can be formulated in the $\RR ^{d_T}$ target subspace. To reduce the 
computational effort, ESA proposes to evaluate the similarity between the target aligned source 
samples and the target subspace projected data by using directly their Euclidean distance \cite{Fernando2014}:
\begin{equation}
\Theta(\xxs,\xxt) = ||\xxs X_a - \xxt X_T ||_2~.
\label{eq:distance}
\end{equation}
The cross-validation procedure described to define the best $d$ for SA becomes very slow and 
tedious when working with data represented by high dimensional features. Moreover, it is unlikely
to provide reliable results in cases where some source classes have an extremely limited number of annotated 
samples. When starting from a rich and reliable representation, one desideratum is to keep 
its strength and retain the sample local neighborhood after dimensionality reduction. With 
this purpose, ESA chooses the domain intrinsic dimensionality obtained through the method 
presented in \cite{Levina2004}. The \emph{Maximum Likelihood Estimate} (MLE) of the 
dimensionality for each data point is calculated and its average is used as the intrinsic 
dimensionality of the corresponding domain \cite{Fernando2014}. The two domains are considered 
separately, which implies $d_S\neq d_T$~. For more details, we refer to \cite{Fernando2014}.

\section{Experiments}
\label{sec:expers}

In this section we provide a detailed experimental analysis on the task of location 
recognition over large time lags using the new LTLL dataset introduced in section~\ref{sec:chall}. 

In the first part of the experiments, we use an image classification framework to evaluate 
different feature detectors, feature descriptors and image representations (section \ref{sec:expfeat}). 
Moreover, we investigate the advantages of using existing domain adaptation methods 
for the considered location recognition problem (section \ref{sec:expdim}). 
All these tests are done using a Nearest Neighbor (NN) classifier. 
Given all the modern training images (source), each labeled with one of the 25 locations, we 
annotate a test ancient picture (target) with the location of the closest/most similar 
modern image. We use the standard Euclidean distance to evaluate the sample similarity unless 
specified otherwise, and equations (\ref{eq:gfk}), (\ref{eq:sim}), (\ref{eq:distance}) when 
applying the corresponding domain adaptation methods. The final performance is always 
evaluated by the multi-class classification accuracy obtained over the full set of old photographs. For this we calculate the percentage of correctly classified images over the full test images.

In the last part of our analysis, we study the task of cross-domain location retrieval and give details about the
application of Extended Subspace Alignment (ESA) with relevance feedback (section \ref{sec:retr}). In this case we consider per-class average precision and take the mean average precision over all classes to obtain mAP.
Several historical query images are accumulated together with their corresponding retrieved modern images. 
We show that by applying domain adaptation over them it is possible to learn a domain-invariant representation 
that provides a significant improvement in the mean average precision results.

\subsection{Seeking The Best Image Representation}
\label{sec:expfeat} 

We start our experimental analysis by establishing which is the best image representation for 
the task of location recognition over large time lags, focusing on those that have been proposed 
as robust to large appearance changes. Most of them are obtained by the combination of local 
descriptors extracted from detected keypoints.

\subsubsection{Setup}
\label{sec:expfeat-setup} 

\noindent
We consider the following

\paragraph{Detectors} Among the existing detectors we test the Difference of Gaussians 
(\textbf{DoG} \cite{marr80}), the Hessian Affine (\textbf{HA}, using the efficient 
implementation proposed in~\cite{Perdoch2009}), and a standard dense sampling strategy (\textbf{Dense}).

\paragraph{Descriptors} As descriptors we consider root-SIFT (\textbf{rSIFT}, \cite{Arandjelovic2012})
and Local Intensity Order Pattern (\textbf{LIOP}, \cite{Wang2011a}).

\paragraph{Representation} Each image is represented either through Bag-of-Words (\textbf{BOW}), or Fisher Vectors (\textbf{FV}). 
In both cases the features are square-root and L2 normalized as suggested in \cite{Perronnin2010}. 
$2\times10^5$ randomly sampled descriptors are used to build a 3000 visual word vocabulary with 
k-means, and to train a Gaussian mixture model (GMM). For FV we reduce the dimensionality of 
rSIFT and LIOP to 64 with PCA and we use a GMM with 64 components obtaining a final feature 
vector of dimension 8192.

\vspace{3mm}
\noindent
We also evaluate features that have pre-defined detector-descriptor pairs.

\paragraph{Self Similarity (\textbf{Self-Sym} \cite{Shechtman2007}) and Symmetry Features 
(\textbf{Sym-Feat}, \cite{Hauagge2012})} 
We follow the same procedure described before to reduce the Self-Similarity descriptor 
dimension to 32 and combine it with a GMM model with 128 components, maintaining the final 
FV dimensionality of 8192. 

\paragraph{Edge Foci detector and Binary Coherent Edge descriptor (\textbf{Edge-Foci+BiCE}, \cite{Zitnick2010})}
This representation is described as robust not only to illumination and pose changes, but 
also to intra-category appearance variation. BiCE is a binary local descriptor, so using a direct 
image-to-image matching procedure is more natural and meaningful than passing through a BOW vocabulary 
or a GMM model for FV encoding. Two images are matched by using the descriptors Hamming distance 
normalized against the total number of extracted points, and comparing the obtained value with a 
pre-defined threshold\footnote{We tested different threshold values and we present here the best obtained result.}.

\vspace{3mm}
\noindent
Finally, we benchmark the classification results obtained with the described representations against 
the performance of two methods that have been previously applied on cross-domain tasks. One is the 
approach presented in \cite{Shrivastava2011} based on the combination of \textbf{HOG features and 
Exemplar SVM} (\textbf{ESVM}, \cite{Malisiewicz2011}). The other is the \textbf{NBNN classifier} \cite{BoimanCVPR08}, 
considering its cross-domain robustness discussed in \cite{Tommasi2013}. \\

We use ~\emph{Acc. all}~ to indicate the accuracy obtained when all new images are used for training a classifier
with on average eleven samples per location; ~\emph{Acc. one}~ indicates instead the accuracy obtained when a single (random) 
new photograph (per class) is used in training. This last setup is quite challenging due to lack of 
training samples. For it we report the average classification accuracy and its standard deviation over 
100 random repetitions to get statistically meaningful results.

\subsubsection{Analysis}
\label{sec:expfeat-analysis} 

All the recognition results are shown in Table~\ref{tbl:compare-representation}, which is 
divided in three parts. The first two are dedicated respectively to BOW and FV with the 
NN classifier. The last part shows the results obtained with the other considered representations 
and classification methods.

With BOW the best performance is obtained when using rSIFT as descriptor and a dense point extraction 
procedure. The effect of the last one is evident in comparison with the corresponding DoG-rSIFT and 
HA-rSIFT results. Due to the huge difference in the visual appearance of old and new images the 
interest points detected by DoG and HA loose their informative power and it seems better to rely 
on a systematic sampling over the whole image provided by the dense extraction. Moreover, LIOP 
presents very low performance, close to random, which suggests that the relative order of pixel intensities in 
the detected local patches changes significantly across the domains.

The symmetry information coded in the Sym-Feat descriptors seems not preserved when 
passing from modern to old images, inducing low recognition results.
On the other hand, Self-Similarity produces the second best results, showing the importance of 
mining the local geometric layout within each image for cross-domain tasks.

\begin{table}[t!]
\small
\centering
\begin{tabular}{ |@{\hspace{1mm}}c@{\hspace{1mm}}|@{\hspace{1mm}}c@{\hspace{1mm}}|@{\hspace{1mm}}c@{\hspace{1mm}}|@{\hspace{1mm}}c@{\hspace{1mm}}|@{\hspace{1mm}}c@{\hspace{1mm}}|@{\hspace{1mm}}c@{\hspace{1mm}}|}\hline
Detec. 			& Descr.			& Repr. 	& Class. 	&  Acc. one (\%)		&  Acc. all  (\%)\\ \hline
DoG 	 		& rSIFT 			& BOW 		& NN 		&  7.5  $\pm$ 2.4 		& 8.7 \\ \hline
DoG			& LIOP 				& BOW 		& NN 		&  7.3 $\pm$ 3.5 		& 7.7 \\ \hline
Dense  			& rSIFT 			& BOW 		& NN 		&  \textbf{19.9 $\pm$ 3.6} 	& \textbf{34.7} \\ \hline
Dense  			& LIOP 				& BOW 		& NN 		&  6.3$\pm$ 1.8		 	& 4.1	 \\ \hline
HA			& rSIFT 			& BOW 		& NN 		&  11.1 $\pm$ 3.1 		& 17.9 \\ \hline
HA			& LIOP 				& BOW 		& NN		&  4.7	$\pm$ 1.9	 	& 9.2	\\ \hline
\multicolumn{2}{|c|@{\hspace{1mm}}}{Self-Sim}	 	& BOW 		& NN 		&  15.8 $\pm$ 3.3 		& 29.6\\ \hline 
\multicolumn{2}{|c|@{\hspace{1mm}}}{Sym-Feat} 		& BOW 		& NN 		&  6.1 $\pm$ 2.4 		& 8.2\\ \hline \hline
DoG 			& rSIFT 			& FV 		& NN 		&  13.3 $\pm$ 2.2 		& 20.9 \\ \hline
DoG			& LIOP 				& FV 		& NN 		&  9.2  $\pm$ 1.5 		& 16.3 \\ \hline
Dense  			& rSIFT 			& FV 		& NN 		&  22.7 $\pm$ 2.9 		& 30.1 \\ \hline
Dense  			& LIOP 				& FV 		& NN 		&  4.9	$\pm$ 1.6	 	& 7.7	 \\ \hline
HA			& rSIFT 			& FV 		& NN		&  \textbf{31.3 $\pm$ 3.5} 	& \textbf{48.5}\\ \hline
HA			& LIOP 				& FV 		& NN		&  4.1	$\pm$ 1.5		& 4.6	\\ \hline
\multicolumn{2}{|c|@{\hspace{1mm}}}{Self-Sim	} 	& FV 		& NN 		&  17.4 $\pm$ 2.8 		& 33.7 \\ \hline 
\multicolumn{2}{|c|@{\hspace{1mm}}}{Sym-Feat} 		& FV 		& NN 		&  14.0 $\pm$ 2.5 		& 26.0 \\ \hline \hline 
Edge-Foci 		& BiCE 		& \multicolumn{2}{c|@{\hspace{1mm}}}{Matching}  &   10.7 $\pm$ 2.6  		& 18.7 
\\ \hline
\multicolumn{3}{|c|@{\hspace{1mm}}}{HOG} 				& ESVM 		&   15.9 $\pm$ 3.5 		& 31.4\\ \hline
HA	& rSIFT 		& FV 					& ESVM 		&   \textbf{28.0 $\pm$ 3.4} 	& \textbf{44.6}\\ \hline
HA	& rSIFT 		& \multicolumn{2}{c|@{\hspace{1mm}}}{NBNN}&   4.7 $\pm$ 1.0 	& 7.1\\ \hline
\end{tabular} 
\caption{Comparison of detectors, descriptors, and image representations. We report the recognition rate results 
over the target (ancient) images  in case of a single source (modern) sample per location (Acc. one),
and when considering the full source set (Acc. all).}
\label{tbl:compare-representation}
\end{table}

The recognition rates obtained with FV are better on average than the corresponding ones based on BOW. 
The trend among the different detector-descriptor cases is analogous to what we discussed before, except 
that the HA detector appears able to complement FV better than dense sampling, leading to the highest performance. 
The disappointing results obtained with Edge-Foci+BiCE indicate that this approach is clearly not suitable 
for the task at hand.

\begin{figure*}[t!]
\small 
\scalebox{0.98}{
 \begin{tabular}{|@{\hspace{1mm}}c|@{\hspace{1mm}}c|@{\hspace{1mm}}c|@{\hspace{1mm}}c|@{\hspace{1mm}}c|@{\hspace{1mm}}c|@{\hspace{1mm}}c|}
 \hline
Test Image & DOG-rSIFT-BOW & Dense-rSIFT-BOW & Self Similarity-FV & HOG-ESVM & HA-rSIFT-FV & ESA\\\hline
 \includegraphics[width=0.12\textwidth]{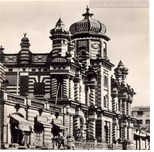} &
 \fcolorbox{red}{red}{\includegraphics[width=0.12\textwidth]{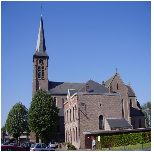}} &
 \fcolorbox{green}{green}{\includegraphics[width=0.12\textwidth]{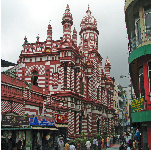}} &
 \fcolorbox{green}{green}{\includegraphics[width=0.12\textwidth]{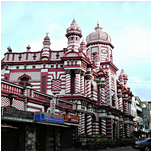}} &
 \fcolorbox{red}{red}{\includegraphics[width=0.12\textwidth]{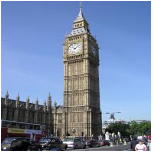}} &
 \fcolorbox{green}{green}{\includegraphics[width=0.12\textwidth]{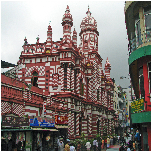}} &
 \fcolorbox{green}{green}{\includegraphics[width=0.12\textwidth]{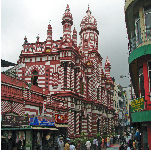}} \\ \hline
 
 \includegraphics[width=0.12\textwidth]{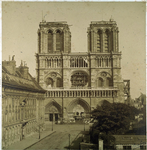} &
 \fcolorbox{red}{red}{\includegraphics[width=0.12\textwidth]{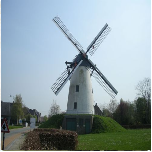}} &
 \fcolorbox{green}{green}{\includegraphics[width=0.12\textwidth]{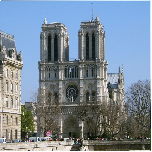}} &
 \fcolorbox{green}{green}{\includegraphics[width=0.12\textwidth]{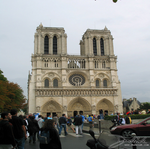}} &
 \fcolorbox{red}{red}{\includegraphics[width=0.12\textwidth]{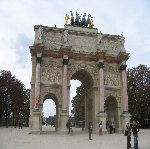}} &
 \fcolorbox{red}{red}{\includegraphics[width=0.12\textwidth]{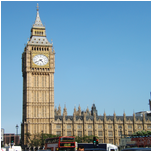}} &
 \fcolorbox{green}{green}{\includegraphics[width=0.12\textwidth]{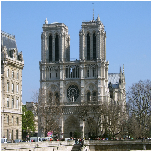}} \\ \hline

 \includegraphics[width=0.12\textwidth]{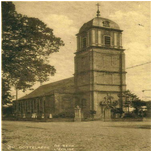} &
 \fcolorbox{red}{red}{\includegraphics[width=0.12\textwidth]{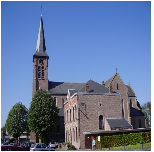}} &
 \fcolorbox{red}{red}{\includegraphics[width=0.12\textwidth]{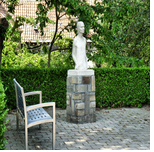}} &
 \fcolorbox{red}{red}{\includegraphics[width=0.12\textwidth]{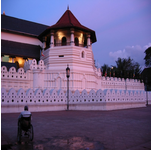}} &
 \fcolorbox{red}{red}{\includegraphics[width=0.12\textwidth]{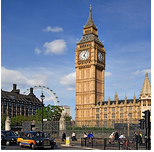}} &
 \fcolorbox{green}{green}{\includegraphics[width=0.12\textwidth]{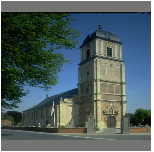}} &
 \fcolorbox{green}{green}{\includegraphics[width=0.12\textwidth]{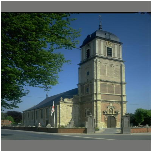}} \\ \hline
 
 \includegraphics[width=0.12\textwidth]{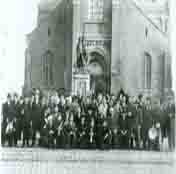} &
 \fcolorbox{red}{red}{\includegraphics[width=0.12\textwidth]{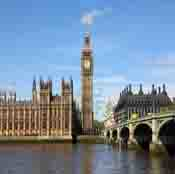}} &
 \fcolorbox{red}{red}{\includegraphics[width=0.12\textwidth]{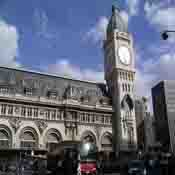}} &
 \fcolorbox{red}{red}{\includegraphics[width=0.12\textwidth]{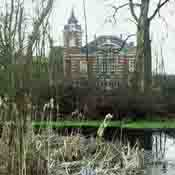}} &
 \fcolorbox{red}{red}{\includegraphics[width=0.12\textwidth]{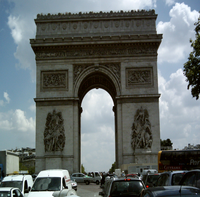}} &
 \fcolorbox{green}{green}{\includegraphics[width=0.12\textwidth]{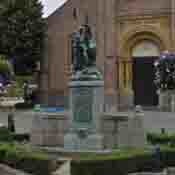}} &
 \fcolorbox{green}{green}{\includegraphics[width=0.12\textwidth]{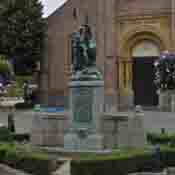}} \\ \hline
 
 \includegraphics[width=0.12\textwidth]{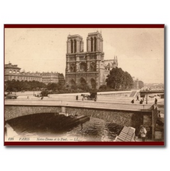} &
 \fcolorbox{red}{red}{\includegraphics[width=0.12\textwidth]{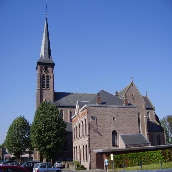}} &
 \fcolorbox{red}{red}{\includegraphics[width=0.12\textwidth]{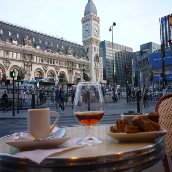}} &
 \fcolorbox{red}{red}{\includegraphics[width=0.12\textwidth]{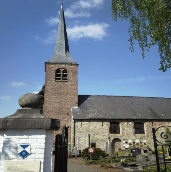}} &
 \fcolorbox{red}{red}{\includegraphics[width=0.12\textwidth]{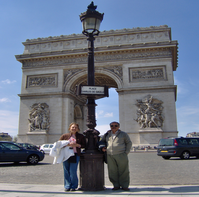}} &
 \fcolorbox{red}{red}{\includegraphics[width=0.12\textwidth]{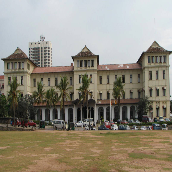}} &
 \fcolorbox{green}{green}{\includegraphics[width=0.12\textwidth]{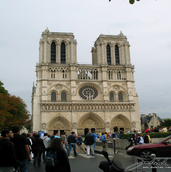}}\\\hline 
 
 \includegraphics[width=0.12\textwidth]{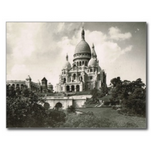} &
 \fcolorbox{red}{red}{\includegraphics[width=0.12\textwidth]{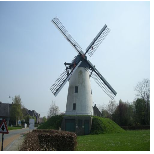}} &
 \fcolorbox{red}{red}{\includegraphics[width=0.12\textwidth]{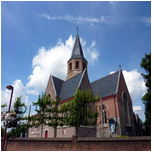}} &
 \fcolorbox{red}{red}{\includegraphics[width=0.12\textwidth]{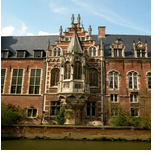}} &
 \fcolorbox{red}{red}{\includegraphics[width=0.12\textwidth]{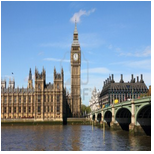}} &
 \fcolorbox{red}{red}{\includegraphics[width=0.12\textwidth]{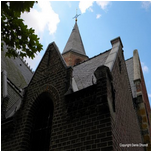}} &
 \fcolorbox{green}{green}{\includegraphics[width=0.12\textwidth]{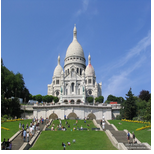}} \\ \hline

 \includegraphics[width=0.12\textwidth]{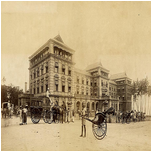} &
 \fcolorbox{red}{red}{\includegraphics[width=0.12\textwidth]{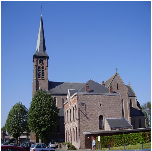}} &
 \fcolorbox{red}{red}{\includegraphics[width=0.12\textwidth]{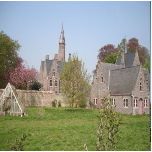}} &
 \fcolorbox{red}{red}{\includegraphics[width=0.12\textwidth]{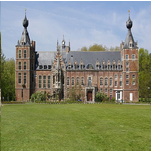}} &
 \fcolorbox{red}{red}{\includegraphics[width=0.12\textwidth]{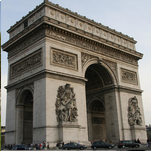}} &
 \fcolorbox{red}{red}{\includegraphics[width=0.12\textwidth]{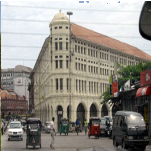}} &
 \fcolorbox{red}{red}{\includegraphics[width=0.12\textwidth]{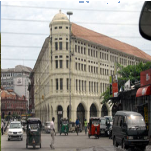}} \\ \hline
 \end{tabular}
}
 \caption{ Examples of the results obtained with different feature representations and with ESA. Given the target test image in the first column, we show here the most similar source images. Red colour indicates wrongly classified instance whereas green indicates correctly classified instance. In the fifth and sixth rows only ESA correctly recognizes Notre Dame and Sacre Coeur. The last row shows a failure for all the methods. By comparing the columns it is visible that different features capture different levels of similarity with the query image and that HOG-ESVM mostly focus on the scene alignment.}
\label{tbl:visualexamples}
\end{figure*}

The combination of HOG features and ESVM present a low performance: as evident in the examples shown in
Figure \ref{tbl:visualexamples}, the HOG features mostly focus 
on the scene alignment, regardless of the specific depicted location. As a variant we also combine 
ESVM with HA-rSIFT-FV and the improved results underline 
the importance of the feature representation. Still, compared to a simple NN classifier, ESVM needs 
a set of extra negative samples besides the choice of learning parameters (\ie tuning the $C$ value),
and does not yield better results. 
Finally the performance of NBNN is almost random, indicating that for the considered task, the image-to-class 
paradigm is not strong enough to overcome the difference among local descriptors in the train and test set.

Overall the combination of HA detector, rSIFT descriptor and FV encoding produces the best results and we 
will use this representation for all the following experiments.

\begin{figure}[htb]
\centering
\includegraphics[width=0.68\linewidth]{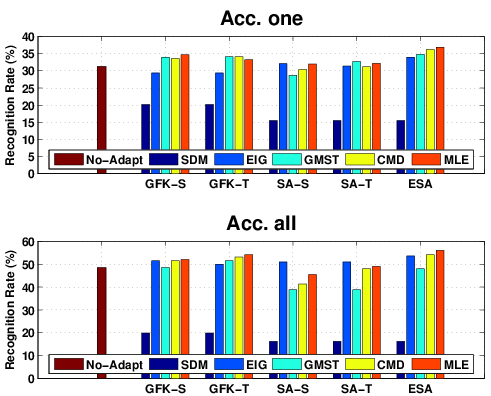}
\caption{Nearest Neighbor classification results of several domain adaptation approaches 
(indicated in the x-axis) when changing the dimensionality estimation method (indicated 
in the legend). No-Adapt corresponds to using HA-rSIFT-FV representation without adaptation. 
-S and -T indicate that the dimensionality of the subspace was estimated on the source or on 
the target domain. For SDM, GFK-S$=$GFK-T and SA-S$=$SA-T. The title of the plot indicates 
that the results were obtained respectively with one sample per location (Acc. one) or 
considering the full source set (Acc. all) of modern images.}
\label{fig:dim}
\end{figure}

\subsection{Domain Adaptation and Subspace Dimensionality}
\label{sec:expdim}

We investigate here the value of domain adaptation in closing the gap between historical and 
modern images. We test the adaptive methods \textbf{GFK}, \textbf{SA} and \textbf{ESA}, 
comparing \textbf{SDM} and \textbf{MLE} against other dimensionality estimation techniques, namely

\begin{description}
\item[EIG:] the eigenvalue-based estimation is the standard solution used in the literature 
for which we choose the dimensionality that retains 99\% of the data variance.
\item[GMST:] the geodesic minimum spanning tree method~\cite{Costa2003} embeds the data in 
a geodesic graph and prunes it to obtain the graph spanning over all the samples with the 
minimum total geodesic length.
\item[CDM:] the correlation dimension technique was proposed in \cite{Decoster1991} to 
approximate the fractal dimension of a dataset. 
\end{description}
Note that the output of SDM is a single subspace dimensionality value for both the 
domains while all the other methods provide two different values, one for each domain. 
We also remark that subspace learning is an unsupervised process, thus all the available 
samples can be used regardless of the availability of their class labels. We 
adopt the standard framework used in previous domain adaptation literature both
for the adaptive and classification process. All modern training images are used to learn 
the source subspace $X_S$ and all ancient testing images are used to learn target subspace $X_T$. 
We then rely on the labels of the source modern images (all or a subset depending on the experiment) 
to annotate the unlabeled test ancient photos. We report the classification accuracies in 
Figure \ref{fig:dim}.

From the histogram bars it can be immediately noticed that all the domain adaptation 
methods in combination with SDM produce worse results than No-Adapt which corresponds 
to using HA+rSIFT+FV and NN without adaptation (which we also reported in Table 
\ref{tbl:compare-representation}). This outcome is not so surprising if we consider 
that, from an original space dimensionality of 8192, the samples are projected to a 
subspace of dimension 16. All the other dimensionality estimation approaches provide 
higher values, for example  EIG=199, GMST=49, CDM=56 and MLE=95 respectively. Even-though EIG a is simple technique, 
the classification accuracy is quite sensitive to the chosen energy percentages 
(99\% in our experiments). Finally,  MLE produces on average the best results with 
respect to all the other
dimensionality estimation techniques.  

\begin{figure}[t!]
\centering
\includegraphics[width=0.65\linewidth]{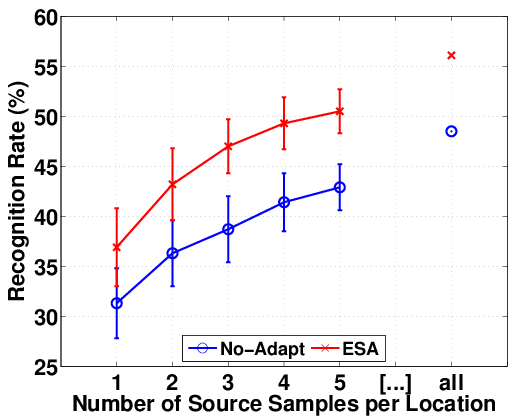}
\caption{Nearest Neighbor classification performance obtained when changing the number 
of source samples per location. The  results showed for 1 and ``all'' corresponds to what 
already shown in Figure \ref{fig:dim} for ESA-MLE.}
\label{fig:changesamples}
\end{figure}

When comparing the domain adaptation methods, we can see that ESA improves over all the 
other approaches. We also test ESA with MLE when varying the number of classifier 
training images between one and five: Figure~\ref{fig:changesamples} shows that even in 
the case of a reduced amount of labeled modern images this approach consistently 
improves over non adaptive classification. 

Finally, to put our results in a wider perspective we add a further benchmark against 
the state of the art deep learning method. In the absence of large amount of training data,  
re-training a CNN network is prone to overfitting  \cite{oneshot}, and fine-tuning the last 
layers of an existing network does not converge, not showing any meaningful learning.
Thus we exploit directly the activation values of a pre-trained network as feature 
representation, namely DeCAF \cite{razavian-arXiv}.
The results are reported in Table~\ref{tab:bestresults} 
together with what was originally achieved without adaptation. 
We notice that ESA applied over FV outperforms what obtained 
with the DeCAF features \cite{Donahue2014}. However, when ESA is applied over DeCAF features, recognition rate obtained with one training sample (Acc. one (\%)) seems to outperforms $\textit{HA-rSIFT-FV} + ESA$.
But when all training samples are used,  $\textit{HA-rSIFT-FV} + ESA$ outperforms $\textit{DeCAF} + ESA$.
We conclude that in the task of 
location recognition over large time lags domain adaptation has a relevant impact 
with a particular advantage provided by  ESA~\cite{Fernando2014} over the other tested approaches.

\begin{table}[t!]
\centering
\begin{tabular}{ | c | c | c |}\hline
Method & Acc. one (\%) & Acc. all (\%) \\ \hline
DeCAF & 36.3 $\pm$ 3.3 & 49.1 \\ \hline 
\textit{HA-rSIFT-FV}  & 31.3 $\pm$ 3.5 & 48.5 \\ \hline 
\textit{HA-rSIFT-FV} + ESA & 36.9 $\pm$ 3.8 & \textbf{56.1} \\ \hline 
\textit{DeCAF} + ESA & \textbf{39.3 $\pm$ 2.7} & 49.0 \\ \hline 
\end{tabular}
\caption{Classification rate obtained with different methods. The last row reports the 
best non-adaptive results of Table \ref{tbl:compare-representation}.}
\label{tab:bestresults}
\end{table}

\subsection{Cross-Domain Location Retrieval}
\label{sec:retr}

In this section we introduce the task of cross-domain location retrieval. 
Given a query old image showing a certain location, the goal is to retrieve modern images which depict the
same location from a database (archive) consisting of few relevant 
images and large number of non-relevant images. Typical image retrieval databases  contain 
$10^4-10^6$ or more samples. To replicate this setting we enlarge our LTLL database
by using images from the Oxford-building 105K database~\cite{Arandjelovic2012}
obtaining a retrieval problem with 225 ancient query images
and a modern image archive with 275 relevant images and 105K distractor images.

As an initial check, we adopt what is considered as best practice in standard instance 
retrieval~\cite{Philbin07,Arandjelovic2012}. We use an image representation obtained by
combining the Hessian Affine detector~\cite{Perdoch2009} with the root-SIFT~\cite{Arandjelovic2012} 
descriptor and BOW with a dictionary size of $[10^4, 10^5, 10^6]$ created through an 
approximate k-means~\cite{Philbin07} and we use the tf-idf scheme. The performance 
obtained in this way is lower than what can be achieved with Fisher Vectors 
(see Table~\ref{tab:retrive.base}).
A similar behavior can be 
observed with other interest point detectors, confirming what we already discussed 
before in section ~\ref{sec:expfeat}. Motivated by the effectiveness of ESA 
to overcome the visual variability induced by large time lags in classification, 
we evaluate its extension to cross-domain location retrieval in the next section.

\begin{table}[htb]
\centering
\begin{tabular}{ | c | c |}\hline
Method & mAP \\ \hline
BOW - 10K &  0.123\\ \hline
BOW - 100K &  0.122\\ \hline
BOW - 1M &  0.086\\ \hline
Fisher Vectors &  0.164\\ \hline
\end{tabular}
\caption{Comparison of BOW and Fisher Vectors (FV parameters as in section~\ref{sec:expfeat}) 
on cross domain location retrieval task 
using the LTLL dataset and the Oxford-building 105K dataset as distractors. 
Old photographs are used as query images and the objective is to 
retrieve new images of the same location depicted in the query image.}
\label{tab:retrive.base}
\end{table}

\subsubsection{Interactive Cross-Domain Retrieval With Domain Adaptation}
\label{sec:retr-dom}

Using domain adaptation in an instance retrieval setting turns out to be quite challenging. The 
reason is that domain adaptation relies on the samples of both the domains to learn and recompose 
the domain shift, but in image retrieval the query (target) samples are not available beforehand, 
while the source data (i.e. the subset of the database corresponding to relevant locations) can 
be identified only as more and more queries are issued.
To overcome this lack of information we relax the problem and make the retrieval process interactive. 
The idea is to ask a user to select three relevant images from the retrieved result set of each query. 
By doing that we are able to collect some query images (old photographs or the target domain) and new 
relevant images (the source domain images). Finally, by using these collected samples we can estimate the 
subspaces of respective domains and use them to perform adaptation by learning 
the subspace alignment matrix $M$ which is then used over new query images.

For the described process it is necessary to control the source and target sample cardinality: we 
need a minimum number of relevance feedback samples and queries to learn a full rank transformation 
matrix. We indicate with $n_S^k$ the number of collected source images obtained with the feedback 
mechanism at round $k$, and with $n_T^k$ the corresponding number of target query images. The 
respective subspace intrinsic dimensionalities $\widehat{d}_S$ and $\widehat{d}_T$ can be calculated 
by using 15 distinct images for each of the two domains: this amount of samples allows to evaluate 
$~100$ pairwise distances and provides enough information to set the local neighborhood of each sample 
for MLE \cite{Fernando2014}. The matrix $M$ is then learned at the first iteration $k=k^*$ which 
satisfies the conditions  $n^{k^*}_S>\widehat{d}_S$ and  $n^{k^*}_T>\widehat{d}_T$. 
\begin{figure}[tb!]
\centering
\includegraphics[width=0.85\linewidth]{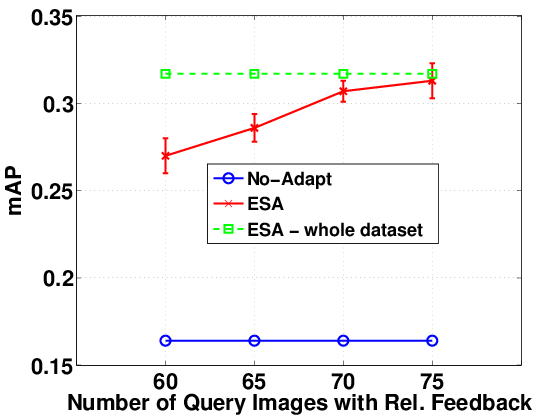}
\caption{Retrieval results obtained when changing the number of query images. In this experiment 
the modern images are used as the reference database together with $10^5$ distractors, while the 
old images are the queries. ``No-Adapt'' corresponds to the result obtained by using HA-rSIFT-FV 
without any adaptation. ``ESA-whole dataset'' refers to the result that can be obtained when 
the transformation matrix $M$ is learned over the full set of old and new images of the 25 
locations in our dataset. ``ESA'' indicates the interactive cross-domain retrieval method.
We refer to the text for further details.}
\label{fig:retr}
\end{figure}
For our target task $\widehat{d}_T=60$ and the source task $\widehat{d}_S=95$, so we collect $60$ distinct queries and $180$ 
feedbacks amounting to about 90-115 distinct modern images. 

After the subspace alignment step over those data we also use PCA whitening \cite{Perronnin2010} 
with the eigenvalues obtained from the query 
images. We repeat this experiment 10 times and we report the obtained mean average precision in 
Figure \ref{fig:retr}, together with the results obtained when increasing the number of query images. 
The plot shows that ESA outperforms the non adaptive solution and with $75$ query samples it reaches 
almost the same results that would have been obtained by learning the transformation matrix $M$ over 
our whole dataset (i.e. the same $M$ used in the classification experiments). We also compare the 
obtained results with a na\"{\i}ve baseline method which exploits directly the similarity among 
the query images. Given a query sample we can first search the most similar image among the 
accumulated historical pictures and then use the associated modern feedback images to search 
in the modern archive. This procedure gives a mAp of $0.201\pm0.023$, which is still lower than 
what we obtained with ESA ($0.313\pm0.010$).

Apart from being effective in the retrieval setting as shown, ESA makes the use of Fisher Vectors 
time and memory efficient since it operates in the low dimensional target space. In our experiments 
we need about 350Mb of RAM for 100K images and a single query is executed in less than 0.03 seconds 
using a single core of 2.8GHz. The matrix $M$ can be learned in a few seconds, which allows ESA domain 
adaptation approach to be applied also in an online setup.

\section{Conclusion}
\label{sec:concl}

In this paper we introduced the task of recognizing the location depicted in an old 
photograph using modern digital images. We presented a dataset spanning over 25 
locations and more than one century and we analyzed several representations looking 
for the most robust to the variability induced by color degradation and different 
image acquisition processes. Our experimental evaluation has shown that Hessian 
Affine detector~\cite{Mikolajczyk2005,Perdoch2009} and root-SIFT~\cite{Arandjelovic2012} 
in combination with Fisher Vectors~\cite{Perronnin2010} are more suitable for the task at 
hand than other detector-descriptor pairs originally introduced to cope with non-linear 
intensity changes~\cite{Hauagge2012,Zitnick2010}.

The difference in visual appearance among old and new images causes a domain shift at image 
descriptor level. Consequently, we obtain poor recognition performance for bag-of-words, 
descriptor matching approaches and NBNN. To overcome this problem we investigated the use 
of domain adaptation methods. Our analysis demonstrated that among different
subspace adaptive learning approaches the Extended Subspace Alignment method~\cite{Fernando2014}
provides the best results and shows a significant advantage in recognition over 
non-adaptive strategies (from $48.5\%$ to $56.1\%$) and state-of-the-art CNN features~\cite{razavian-arXiv} ($49.1\%$).

Finally we proposed and analyzed the task of cross-domain location retrieval.
We proposed a strategy to interactively use domain adaptation and showed the gain in performance
provided by ESA also in this setting (from $0.201 $ to $0.313$ mAP).

Our work presents several cues that indicate good directions for future research.
We believe that the LTLL dataset introduced in this paper is a good testbed to evaluate the 
practical usefulness of existing domain adaptation methods. We plan to extend the collection 
and to investigate how adaptive methods scale in case of more samples and an 
increasing number of classes/locations. Indeed the application of domain adaptation on 
large datasets and the effect on their speed/complexity and accuracy have not been
extensively studied yet.
The proposed dataset may also influence the location recognition community to seek novel
image representations that are not susceptible to distribution mismatch due to large
time lags. 
Moreover our analysis suggests that there is a great necessity of new learning
algorithms able to overcome the domain-shift issue in the cross-domain image
retrieval setting. On one side the presented study paves the way for online-interactive
domain adaptation systems, on the other it may inspire new instance
retrieval methods and paradigms~\cite{Fernando2013a,Arandjelovic2012a}.

\textbf{Acknowledgements} :The authors acknowledge the support of the EC FP7 project AXES and iMinds Impact project Beeldcanon.

\end{document}